\documentclass[sigconf]{acmart}
\usepackage{balance}
\usepackage{multirow}
\usepackage{amsmath,amssymb,amsfonts}
\usepackage{bm}
\usepackage{subfigure}

\AtBeginDocument{%
  \providecommand\BibTeX{{%
    \normalfont B\kern-0.5em{\scshape i\kern-0.25em b}\kern-0.8em\TeX}}}

\makeatletter 
\renewcommand{\@thesubfigure}{\hskip\subfiglabelskip}
\makeatother

\copyrightyear{2020}
\acmYear{2020}
\setcopyright{acmcopyright}\acmConference[MM '20]{Proceedings of the 28th ACM International Conference on Multimedia}{October 12--16, 2020}{Seattle, WA, USA}
\acmBooktitle{Proceedings of the 28th ACM International Conference on Multimedia (MM '20), October 12--16, 2020, Seattle, WA, USA}
\acmPrice{15.00}
\acmDOI{10.1145/3394171.3413583}
\acmISBN{978-1-4503-7988-5/20/10}

\acmSubmissionID{1068}

\begin{document}

\title{Dual Semantic Fusion Network for Video Object Detection}



\author{Lijian Lin}
\authornote{This work is done while interning at Applied Research Center (ARC), Tencent PCG.}
\authornote{Equal contribution.}
\affiliation{%
  \institution{\small{Fujian Key Laboratory of Sensing and Computing for Smart City, School of Informatics, Xiamen University, Xiamen, China.}}
 }
 \email{ljlin@stu.xmu.edu.cn}
 
 \author{Haosheng Chen}
\authornotemark[2]
\affiliation{%
  \institution{\small{Fujian Key Laboratory of Sensing and Computing for Smart City, School of Informatics, Xiamen University, Xiamen, China.}}
 }
 \email{haoshengchen@stu.xmu.edu.cn}

\author{Honglun Zhang}
 \affiliation{%
  \institution{\small{Applied Research Center (ARC), Tencent PCG}}
 }
\email{honlanzhang@tencent.com}

\author{Jun Liang}
\affiliation{%
  \institution{\small{Fujian Key Laboratory of Sensing and Computing for Smart City, School of Informatics, Xiamen University, Xiamen, China.}}
 }
\email{Junliang@stu.xmu.edu.cn}

\author{Yu Li, Ying Shan}
 \affiliation{%
  \institution{\small{Applied Research Center (ARC), Tencent PCG}}
 }
\email{{ianyli, yingsshan}@tencent.com}

\author{Hanzi Wang}
\authornote{Corresponding author.}
\affiliation{%
  \institution{\small{Fujian Key Laboratory of Sensing and Computing for Smart City, School of Informatics, Xiamen University, Xiamen, China.}}
 }
\email{hanzi.wang@xmu.edu.cn}

\begin{abstract}
	Video object detection is a tough task due to the deteriorated quality of video sequences captured under complex environments. Currently, this area is dominated by a series of feature enhancement based methods, which distill beneficial semantic information from multiple frames and generate enhanced features through fusing the distilled information. However, the distillation and fusion operations are usually performed at either frame level or instance level with external guidance using additional information, such as optical flow and feature memory. In this work, we propose a dual semantic fusion network (abbreviated as DSFNet) to fully exploit both frame-level and instance-level semantics in a unified fusion framework without external guidance. Moreover, we introduce a geometric similarity measure into the fusion process to alleviate the influence of information distortion caused by noise. As a result, the proposed DSFNet can generate more robust features through the multi-granularity fusion and avoid being affected by the instability of external guidance. To evaluate the proposed DSFNet, we conduct extensive experiments on the ImageNet VID dataset. Notably, the proposed dual semantic fusion network achieves, to the best of our knowledge, the best performance of 84.1\% mAP among the current state-of-the-art video object detectors with ResNet-101 and 85.4\% mAP with ResNeXt-101 without using any post-processing steps.
\end{abstract}



\maketitle
\section{Introduction}
\label{sec:Introduction}
\begin{figure}[t]
	\begin{center}
		\includegraphics[width=0.90\linewidth]{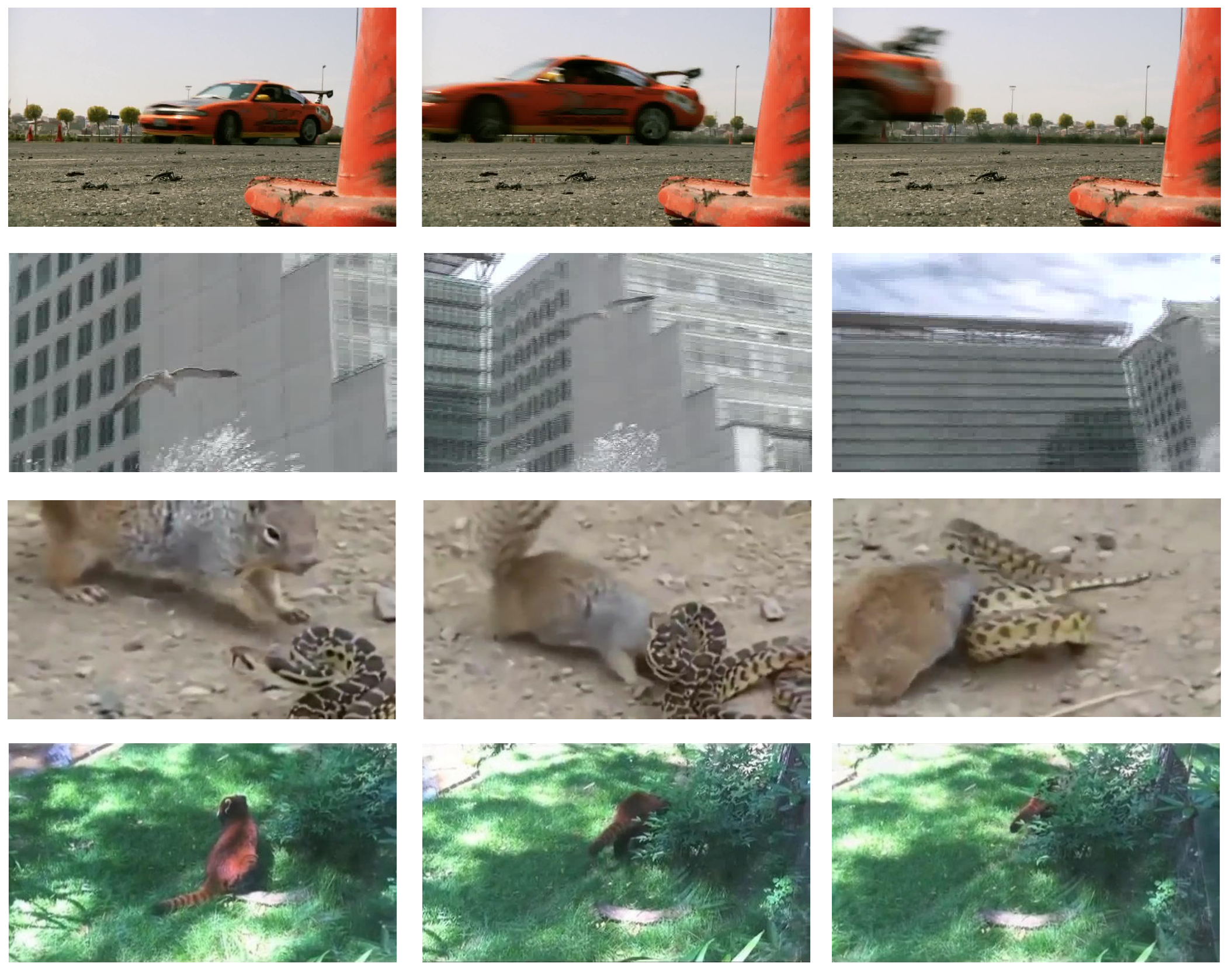}
	\end{center}
	\caption{Some examples of deteriorated video sequences. From the top row to the bottom row, the corresponding challenges are motion blur, video defocus, pose variation and occlusion, respectively.}
	\label{fig:pic1}
\end{figure}

Video object detection aims to detect objects of interest on consecutive video frames, which is a vital task in the multimedia area. Despite the great success achieved by still image detection works, video object detection is still challenging due to the deteriorated video quality caused by motion blur, video defocus, pose variation and occlusion (see Figure \ref{fig:pic1} for some examples). However, along with these challenges, videos inherently contain much richer context information in the spatio-temporal domain compared with individual images, which gives a clear direction of exploiting the rich information in video sequences to improve the performance of video object detection.

Based on the success of single-frame detectors, cutting-edge video object detection works (such as \cite{Fgfa,MM_two_stage2,SELSA,RDN}) tend to consider more than one frame as support frames to leverage the context information of video sequences. Specifically, these works distill useful information from the support frames and then fuse the distilled information into the deteriorated frames to generate enhanced features for robust detection. The distillation and fusion operations are usually performed on either frame-level or instance-level features with external guidance, such as optical flow \cite{Fgfa,THP} or global/local feature memory \cite{deng2019object}. The external guidance is used to measure the similarities among pixels or instances, which are employed to guide the following fusion process. Since the external guidance is usually implemented by using additional deep neural networks, the performance of the guidance based object detection methods cannot be assured by themselves and they may suffer from the instability of the external guidance. In particular, false positive estimations introduced by the external guidance are fatal for the guidance based methods and they are more likely to cause the failure of detection.

Considering that there are frame-level and instance-level semantic information that can be extracted by the dominated two-stage detection framework before and after the region proposal network (RPN), there is a natural desire to perform the distillation and fusion operations at both levels. Consequently, in this paper, we present a Dual Semantic Fusion Network (called DSFNet) to exploit both frame-level and instance-level semantic information in video sequences. Moreover, we propose a geometric similarity measure to measure the geometric similarities among object instances. Then, the geometric similarities are used to cooperate with its corresponding appearance information to mitigate the information distortion problem caused by noise during the dual semantic fusion procedure. Compared with the current one-stage feature enhancement methods, the proposed DSFNet can generate more robust features through the multi-granularity fusion. 

Different from the existing guidance based methods, we argue that the proposed dual semantic fusion network can be learned through a unified framework in a fully end-to-end manner, even if there is no external guidance for the fusion. The reason is that the frame-level fusion in DSFNet can provide rich but object-agnostic information, which consists of relatively low-level semantics. On the contrary, the instance-level fusion in DSFNet can distill object-specific but limited information, which includes relatively high-level semantics, as the complementary cue. Through the combination of the frame-level and instance-level semantic fusions, the distilled information from one level becomes the internal guidance of the other level in DSFNet. In addition, since we do not use any external guidance in our network, the proposed DSFNet is self-contained and it does not need to rely on the precision and reliability of the corresponding external guidance.

In summary, we make the following contributions in this paper: 

\begin{itemize}
	\item We present a dual semantic fusion network, which performs a multi-granularity semantic fusion at both frame level and instance level in a unified framework and then generates enhanced features for video object detection. 
	\item We introduce a geometric similarity measure into the proposed dual semantic fusion network along with the widely used appearance similarity measure to alleviate the information distortion caused by noise during the fusion process.
	\item We explain the video object detection process from a novel information theory perspective and then give a detailed analysis to show the effectiveness of the proposed dual semantic fusion network.
\end{itemize}

We evaluate the proposed DSFNet on the large scale ImageNet VID dataset \cite{VID}. The experimental results demonstrate the superiority of our DSFNet over several state-of-the-art methods. Especially, DSFNet outperforms its baseline detector by a large margin of 9.4\% mAP and achieves, to the best of our knowledge, the highest mAP of 84.1\%/85.4\% with ResNet-101/ResNeXt-101 when it is compared with the published state-of-the-art video object detection methods  without using additional post-processing steps.


\section{Related Work}
\label{sec:RelatedWork}
In this section, we briefly review several representative still image object detectors and video object detectors.

{\bfseries Still Image Object Detectors.} 
Object detection in still images is one of the fundamental tasks in the multimedia and computer vision communities with a variety of applications (\emph{e.g.}, \cite{MM_app_1, MM_app_2, MM_app_3, MM_app_4, MM_app_5, MM_app_6, TOMM_app_1, TMM_app_1, TMM_2}). State-of-the-art still image object detectors can be roughly classified into two types: two-stage detectors (such as \cite{Rcnn, MM_two_stage1, MM_two_stage3, TMM_two_stage, TMM_two_stage1, TMM_two_stage2, TMM_two_stage3, TMM_two_stage4, cascade-rcnn}), and one-stage detectors (\emph{e.g.}, \cite{Yolo, MM_one_stage1, MM_one_stage2, MM_one_stage3, Yolov4, Cornernet,Fpn}). 

As one of the most representative two-stage detectors, Faster R-CNN \cite{Fasterrcnn} proposes to generate region proposals by using CNNs, and then classifies and refines the generated proposals for object detection. FPN \cite{Fpn} improves Faster R-CNN by designing a feature pyramid network for detecting objects at different scales.
\cite{Relation} proposes to model the similarities among instances to capture the contextual information in a whole image, which yields promising performance on the object detection task.

In contrast, the one-stage detectors directly make predictions with a single detection network. For example, SSD \cite{Ssd} and RetinaNet \cite{Retinanet} place some pre-designed anchor boxes densely over feature maps, and directly classify and refine each anchor box. CenterNet \cite{centernet} and FCOS \cite{Fcos} propose to detect objects in images without designing a set of anchor boxes, and they achieve better performance than most of the anchor based one-stage detectors.

Different from detecting objects in still images, a video sequence contains much richer spatio-temporal information for detection. The rich information can be leveraged to solve the challenging situations (such as occlusion, motion blur, rare poses) when detecting objects in videos. Therefore, in this paper, we propose to exploit the spatio-temporal information in video frames to improve the video object detection performance. Similar to most of the state-of-the-art video object detectors, our proposed DSFNet is also built upon the effective Faster R-CNN framework.


\begin{figure*}
	\centering
	\includegraphics[width=0.90\linewidth]{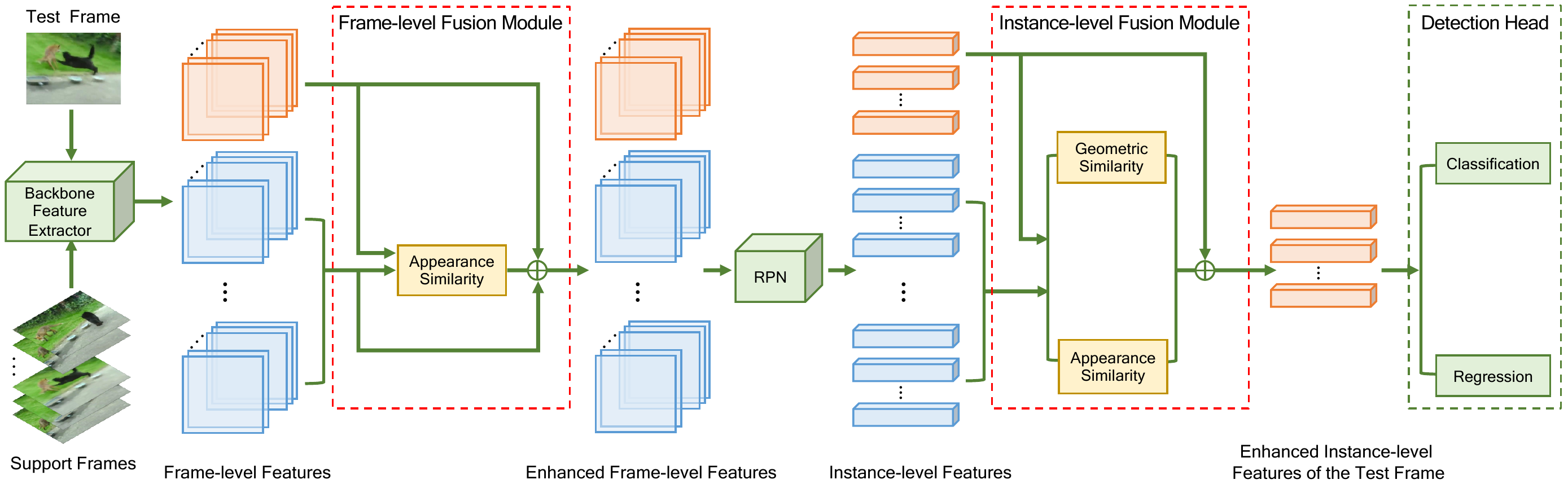}
	\caption{The pipeline of the proposed DSFNet. Given a test frame and several support frames, we first extract their frame-level features. The features marked in orange/blue are the features of the test/support frames. Then, these features are enhanced by fusing them according to their appearance similarities (Eq. (\ref{eq:pixel_relation})). We apply RPN on the enhanced frame-level features to get instance-level features. After that, the instance-level features are enhanced based on their appearance and geometric similarities (Eq. (\ref{eq:instance_relation})). Finally, the enhanced instance-level features of the test frame are fed into the detection head for final detection.} 
	\label{fig:pipeline}
\end{figure*}

{\bfseries Video Object Detectors.}
For the task of video object detection, one of the main challenges is how to improve single-frame detection performance by exploring temporal information of videos. One common solution is to apply post-processing techniques to the predicted bounding boxes obtained by still image detectors \cite{Seqnms, Tcnn, object-link, post_2}. For example, SeqNMS \cite{Seqnms} conducts a sequence-level NMS on the detected bounding boxes, and it uses high-scoring detection results to boost the scores of weaker detection results. T-CNN \cite{Tcnn} adopts optical flow to propagate the predicted detection results to neighboring frames and re-scores the detection results by incorporating an additional object tracker. These post-processing methods can improve the performance of still image object detectors when they are applied to videos. However, the performance of these methods highly relies on their associated image object detectors, and it is difficult for them to correct the errors produced by the associated image object detectors. In contrast, our DSFNet focuses on using the temporal information at feature level rather than at the final bounding box level, and it can be trained in an end-to-end manner without using any post-processing steps.

Recently, state-of-the-art video object detectors tend to fuse features extracted from multiple frames in a video sequence to boost the detection performance. There are mainly two ways to fuse features: frame-level fusion \cite{THP, MM_two_stage2, Associate-LSTm, bottleneck-lstm, memory-guided, STMN} and instance-level fusion \cite{SELSA, RDN, Leverage_long_range}. For the frame-level fusion based methods, they propose to fuse the features extracted from multiple frames at frame level. For example, FGFA \cite{Fgfa} proposes to warp the features from adjacent frames and fuse them to the reference frame according to the optical flow generated by \cite{flownet}. THP \cite{THP} also uses optical flow to propagate the extracted features from keyframes to non-keyframes. STSN \cite{STSN} adopts deformable convolution to align and aggregate features between frames without using optical flow information.

For the instance-level fusion based methods, they fuse the features extracted from object instances for detection. For example, SELSA \cite{SELSA} introduces a semantics aggregation module to fuse the features extracted from object instances and produce enhanced features for detection. RDN \cite{RDN} designs relation distillation networks to measure the relation among object instances and then aggregates them to augment the features of instances for detection. The work in \cite{Leverage_long_range} modifies the non-local block in \cite{non-local} to learn the appearance similarities between the target proposals and the proposals generated from multiple support frames to enrich the target proposal features, which boosts its detection performance. Different from the above-mentioned methods, we propose a dual semantic fusion network, which fuses features on both frame level and instance level. By doing this, the features from both levels are fused to generate enhanced features by utilizing the spatial-temporal information in videos, leading to a better video object detection performance.

\section{Method}
\label{sec:Method}
In this section, we first provide an overview of the proposed DSFNet. Then, we introduce the proposed dual semantic fusion network, which generates enhanced features to perform robust video object detection. Finally, we analyze the proposed DSFNet from an information theory viewpoint.

\subsection{Overview}
The overall video object detection pipeline of the proposed DSFNet is illustrated in Figure \ref{fig:pipeline}. Given a test frame of a video sequence, we first sample a set of support frames from the video sequence and extract the frame-level features of these frames by using the backbone feature extractor. Then, we apply the proposed frame-level semantic fusion module on these features to obtain the corresponding enhanced frame-level features. These enhanced features are fed into the Region Proposal Network (RPN) to generate a set of object instances. We further enhance the features of these instances by using the proposed instance-level semantic fusion module. Finally, we feed the enhanced instance-level features into the detection head for object classification and bounding box regression.

\subsection{Dual Semantic Fusion}
Considering that there are usually deteriorated video frames occurring in the task of video object detection, the main challenge of accurately detecting objects in videos lies in how to leverage the rich information in videos. In this subsection, we describe the proposed dual semantic fusion network, which consists of a frame-level fusion module and an instance-level fusion module. Both fusion modules can enhance the features extracted from individual frames by fusing the rich information in videos.

{\bfseries Frame-level Semantic Fusion.}
Given a test frame in a video sequence, we first sample $n-1$ support frames from the rest of the video sequence. With the $n$ frames, we extract a series of frame-level features $\bm{\mathcal{F}} = \{\bm{F}_1, \bm{F}_2, ..., \bm{F}_{n}\}$, where $\bm{F}_i \in \bm{\mathcal{F}}$ indicates the frame-level feature extracted from the $i$-th input frame. Since each of the frame-level features in $\mathcal{F}$ has $d$ channels, we split the frame-level features in $\mathcal{F}$ into $n*d$ separated channel-wise features $\bm{\mathcal{F}}^c = \{\bm{F}^c_1, \bm{F}^c_2, ..., \bm{F}^c_{n*d}\}$. During the frame-level semantic fusion, inspired by the non-local network in \cite{non-local}, we calculate a similarity matrix $\bm{S}^{F}$ of $\bm{\mathcal{F}}^c$ to represent the appearance similarities among the features in $\bm{\mathcal{F}}^c$. Then, for the $i$-th feature $\bm{F}_i^c$ in $\bm{\mathcal{F}}^c$, we fuse all the features in $\bm{\mathcal{F}}^c$ into $\bm{F}_i^c$ based on $\bm{S}^{F}$ to generate the corresponding $i$-th enhanced feature $\bm{F}_i^e$. Here, we denote the generated enhanced features as $\bm{\mathcal{F}}^e = \{\bm{F}_1^e, \bm{F}_2^e, ..., \bm{F}_{n*d}^e\}$. Specifically, the $i$-th enhanced feature $\bm{F}_i^e \in \bm{\mathcal{F}}^e$ is calculated by the following equation:
\begin{equation}
\label{eq:pixel_agg}
\bm{F}_i^e = \bm{F}_i^c + \sum_{j=1}^{n*d} S_{i,j}^{F} \cdot {\theta}(\bm{F}_{j}^c), i = 1,2,3,...,n*d
\end{equation}
where ${\theta}(\cdot)$ denotes a general transformation function parameterized by fully connected layers. $S_{i,j}^{F} \in \bm{S}^{F}$ means the appearance similarity between $\bm{F}_{i}^c$ and $\bm{F}_{j}^c$, which is calculated as follows:
\begin{equation}
\label{eq:pixel_relation}
S_{i,j}^{F} = \frac{exp{(a_{i,j}})}{\sum_{u=1}^{n*d} exp{( a_{i,u})}}
\end{equation}
where $a_{i,j}$ is the cross product between $\bm{F}_{i}^c$ and $\bm{F}_{j}^c$, and it is formulated as follows:
\begin{equation}
\label{eq:pixel_s}
a_{i,j} = <\phi(\bm{F}_{i}^c), \varphi(\bm{F}_{j}^c)>
\end{equation}
$\phi(\cdot)$ and $\varphi(\cdot)$ are two general transformation functions, which are similar to ${\theta}(\cdot)$. After the fusion, the information contained in the $i$-th feature $\bm{F}_i^c \in \bm{\mathcal{F}}^c$ is propagated to the other features in $\bm{\mathcal{F}}^c$. As a result, each of the enhanced features $\bm{\mathcal{F}}^e$ can distill rich information from the frame-level features of the other frames. 

{\bfseries Instance-level Semantic Fusion.}
For the instance-level semantic fusion, the enhanced features $\bm{\mathcal{F}}^e$ generated by the frame-level semantic fusion module are fed into RPN to generate a set of object instances with the associated bounding boxes $\bm{\mathcal{B}} = \{\bm{B}_1,\bm{B}_2, ..., \bm{B}_{m}\}$. Here $m$ is the number of the generated instances. Each bounding box in $\bm{\mathcal{B}}$ contains the spatial location and the scale information of an instance. Then, a RoI layer is applied on the bounding boxes in $\bm{\mathcal{B}}$ and the enhanced frame-level features in $\bm{\mathcal{F}}^e$ to generate the corresponding RoI features $\bm{\mathcal{Q}} = \{\bm{Q}_1, \bm{Q}_2, ..., \bm{Q}_{m}\}$ of the instances.
After the instance-level semantic fusion, the final enhanced instance-level features $\bm{\mathcal{Q}}^e = \{\bm{Q}_1^e, \bm{Q}_2^e, ..., \bm{Q}_{m}^e\}$ are generated by fusing all the RoI features in $\bm{\mathcal{Q}}$, which is written as follows:
\begin{equation}
\bm{Q}_k^e = \bm{Q}_k + \sum_{l=1}^{m}{S}_{k,l}^{I} \cdot \gamma(\bm{Q}_{l}), k = 1,2,3,...,m
\end{equation}
where $\gamma(\cdot)$ is a general transformation function and ${S}_{k,l}^{I}$ indicates the instance-level similarity between $\bm{Q}_k$ and $\bm{Q}_{l}$.

Since geometric information plays an important role in representing an object as well as the appearance information, for the instance-level fusion, we propose to measure the similarities among the instances not only based on the appearance information contained in $\bm{\mathcal{Q}}$, but also based on the geometric information contained in $\bm{\mathcal{B}}$, which is
\begin{equation}
\label{eq:instance_relation}
{S}_{k,l}^{I} = \frac{exp(z_{k,l} + r_{k,l})}{\sum_{v=1}^{m}exp(z_{k,v} + r_{k,v})}
\end{equation}
where $z_{k,l}$ is the appearance similarity between $\bm{Q}_k$ and $\bm{Q}_{l}$. $r_{k,l}$ indicates the geometric similarity between the $k$-th and $l$-th bounding boxes $\bm{B}_{k}$ and $\bm{B}_{l}$ in $\bm{\mathcal{B}}$. Specifically, $z_{k,l}$ is formulated as:
\begin{equation}
z_{k,l} = <\xi(\bm{Q}_k), \zeta(\bm{Q}_l)>
\end{equation}
where $\xi(\cdot)$ and $\zeta(\cdot)$ are two general transformation functions parameterized by fully connected layers. Since different objects may have similar spatial locations in different frames, the scale information (\emph{i.e.}, the width $w$ and the height $h$) contained in $\bm{\mathcal{B}}$ is more reliable in measuring the geometric similarity than the spatial information. Therefore, we propose to measure the geometric similarity $r_{k,l}$ between $\bm{B}_{k}$ and $\bm{B}_{l}$, as follows:
\begin{equation}
\label{eqgr}
r_{k,l} = \psi(\varrho(log(\frac{w_k}{w_{l}}), log(\frac{h_k}{h_{l}}), log(|\frac{w_k}{h_k} - \frac{w_{l}}{h_{l}}|)))
\end{equation}
$\psi(\cdot)$ indicates a general transformation function, which plays a similar role as $\xi(\cdot)$ and $\zeta(\cdot)$. $\varrho(\cdot)$ is the embedding function used in \cite{Relation}, which embeds the primitive low-dimensional geometric similarity $r_{k,l}$ into a high-dimensional representation for the proposed deep detection network. By exploiting the geometric information and the appearance information, our DSFNet can alleviate the information distortion problem caused by noise during the fusion process. Finally, the enhanced features in $\bm{Q}^e$ that correspond to the test frame are fed to the detection head for the final object detection.

\subsection{An Information Theory Viewpoint}
\label{sec:InformationTheoryViewpoint}
As described in the pioneering study \cite{shwartz2017opening}, the learning process of a deep neural network based object detector can be mathematically analyzed from the perspective of information bottleneck (IB) theory \cite{tishby2000information}, as shown in the state-of-the-art still image object detection work \cite{wang2020bidet}. Similarly, from the IB perspective, learning a deep video object detection network can be considered as a Markov process with a concise Markov chain: 
\begin{equation}
\label{eqMarkov1}
V \rightarrow F \rightarrow O 
\end{equation}
where $V$ is the input variable (\emph{i.e.}, the input video sequence tensor), $F$ means the intermediate variable, which is related to the frame-level features extracted by the backbone network, and $O$ stands for the output variable, which consists of the final predicted object labels and locations. Then, the goal of learning the whole detection network is to minimize the mutual information between the input video tensor $V$ and the frame-level features $F$, and to maximize the mutual information between $F$ and $O$, which is formulated as:
\begin{equation}
\label{eqIB}
\min _{\omega_{b}, \omega_{d}}\{I(V ; F)-\beta I(F ; O)\}
\end{equation}
where $\omega_{b}$ and $\omega_{d}$ are the learnable parameters of the backbone and the detection head, respectively. $\beta$ is a Lagrange multiplier. $I(X, Y)$ is the mutual information between $X$ and $Y$, which is defined as:
\begin{equation}
I(X ; Y)=\mathbb{E}_{\boldsymbol{x} \sim p(\boldsymbol{x})} \mathbb{E}_{\boldsymbol{y} \sim p(\boldsymbol{y} | \boldsymbol{x})} \log \frac{p(\boldsymbol{y} | \boldsymbol{x})}{p(\boldsymbol{y})}
\end{equation}
where $\boldsymbol{x}$ and $\boldsymbol{y}$ are respectively a specific instance in $X$ and the corresponding instance in $Y$. $p(\cdot)$ is the prior distribution, and $\mathbb{E}$ refers to the expectation function. In Eq. (\ref{eqIB}), $I(V ; F)$ is minimized to distill the useful information for the video object detection task from the input variable $V$, while $I(F ; O)$ is maximized to preserve more distilled information for the final detection.

According to the data processing inequality concept in information theory, there is no post-processing that can increase the information contained in the input variable $V$. Specifically, the information contained in the input variable $V$ can not be increased through the given Markov chain, which can be formulated as:
\begin{equation}
\label{eqInequality}
I(V ; F) \geqslant I(V ; O)
\end{equation}
The equality of Eq. (\ref{eqInequality}) can be achieved if and only if $F$ and $O$ contain the same information about $V$, which is impractical due to the high compression in Eq. (\ref{eqIB}). As a consequence, the information contained in $V$ is gradually decreased during the learning process. 

Since most of the state-of-the-art video object detectors and the proposed DSFNet have two stages (\emph{i.e.}, RPN based object proposal generation and RCNN based final prediction) between $F$ and $O$, the Markov chain in Eq. (\ref{eqMarkov1}) can be rewritten as:
\begin{equation}
\label{eqMarkov2}
V \rightarrow F \rightarrow P \rightarrow O 
\end{equation}
where $P$ represents the instance-level features generated by the RoI layer. As a result, following the afore-mentioned data processing inequality rule, we can rewrite Eq. (\ref{eqInequality}) as: 
\begin{equation}
\label{eqInequality2}
I(V ; F) \geqslant I(V ; P) \geqslant I(V ; O)
\end{equation}
According to the IB principle in Eq. (\ref{eqIB}), $I(F ; O)$ should be maximized to preserve more distilled information. Different from still image object detection, for video object detection, the information contained in the support frames of a video sequence can boost the performance of detection on the test frame of the video sequence. Since the information contained in $F$ is gradually decreased in Eq. (\ref{eqInequality2}), and the final prediction is solely based on the information contained in the test frame, the information contained in the support frames should be distilled and fused into the enhanced features of the test frame before and after the RoI layer.

In the proposed DSFNet, the first frame-level fusion is employed before the RoI layer to distill frame-level information from support frames. The distilled frame-level information is then fused into the frame-level features of the test frame. Then, the second fusion is performed at the instance level after the RoI layer. The instance-level fusion, which is a high-level semantic fusion, is based on the object-level similarity. By doing the dual semantic fusion, the information distilled at both the frame level and the instance level is fused into the RoI features of the test video frame for the final object detection. As a result, compared with the current single fusion based methods, the proposed DSFNet can preserve more beneficial information contained in $F$ to perform more accurate video object detection.

\section{Experiments}
\label{sec:Experiments}
In this section, we first introduce the dataset and evaluation protocols for the video object detection task. Then, we present the implementation details of the proposed DSFNet. We also carry out several ablation studies on the ImageNet VID validation set \cite{VID} to verify the effectiveness of the proposed dual semantic fusion network.
Finally, we compare DSFNet with several other state-of-the-art video object detection methods. 

\subsection{Dataset and Evaluation Protocols}
We conduct the experiments on the ImageNet VID dataset \cite{VID}, which is a large scale benchmark for video object detection. The ImageNet VID dataset contains 4,417 video snippets for training and validation. There are 3,862 video snippets in the training set. And the validation set consists of 555 video snippets. 
The frames in these snippets are fully annotated over 30 object categories with bounding boxes. Following the widely adopted protocols in video object detection \cite{Fgfa, RDN, STSN, STMN, SELSA, Tcnn}, we evaluate the proposed DSFNet on the ImageNet VID validation set and use the mAP@IoU=0.5 scores as the evaluation metric. Moreover, as in \cite{Fgfa, SELSA, Leverage_long_range}, all the objects in the ImageNet VID validation set are categorized into three groups (\emph{i.e.}, the slow, medium and fast motion groups), according to their motion speed. We evaluate DSFNet on the objects with different motion groups for better analysis.

Although there are more than a million frames in the ImageNet VID training set, some of these frames are redundant. Thus, the appearance diversity of the objects in the ImageNet VID training set is limited, which makes the training process less effective. Therefore, as in the previous works \cite{Fgfa, RDN, STSN, STMN, SELSA, Tcnn}, we train the proposed DSFNet on the intersection of the ImageNet VID and DET datasets \cite{VID}. The ImageNet DET dataset is a still image detection dataset.

\subsection{Implementation Details}
Next, we will discuss the implementation details of the proposed DSFNet from four aspects, including feature extractor, detection network, dual semantic fusion, and training/inference details.

{\bfseries Feature Extractor.} We use ResNet-101 \cite{Resnet101} or ResNeXt-101-32$\times$4d \cite{Resnext101} as our backbone feature extractor. Following the work in \cite{Fgfa, RDN, SELSA, MANet}, we modify the convolutional stride of the last block of the last stage (i.e, $conv5$ for both ResNet-101 and ResNeXt-101) from 2 to 1. As a result, the total stride of $conv5$ is changed from 32 to 16, which increases the resolution of the extracted feature maps. In addition, we set the dilation rate to 2 in those convolutional layers in $conv5$, where their kernel size is larger than 1, to retain the receptive field of the backbone feature extractor.   

{\bfseries Detection Network.} We adopt Faster R-CNN \cite{Fasterrcnn} as our baseline detection network. RPN is applied to the output of $conv4$. To reduce redundancy, a non-maximum suppression (NMS) with a IoU threshold of 0.7 is adopted and $300$ candidate boxes are generated in each frame during both training and inference phases. A RoI pooling layer is applied to the output of $conv5$ with the generated candidate boxes to extract a series of RoI-pooled features for these boxes. Then, these RoI-pooled features are fed into the detection head for object classification and bounding box regression.

\begin{figure*}[h]
	\centering
	\subfigure[\normalsize{(a) number of test frames} \label{fig:pic31}]{\includegraphics[height = 4.2cm, width = 0.335\linewidth]{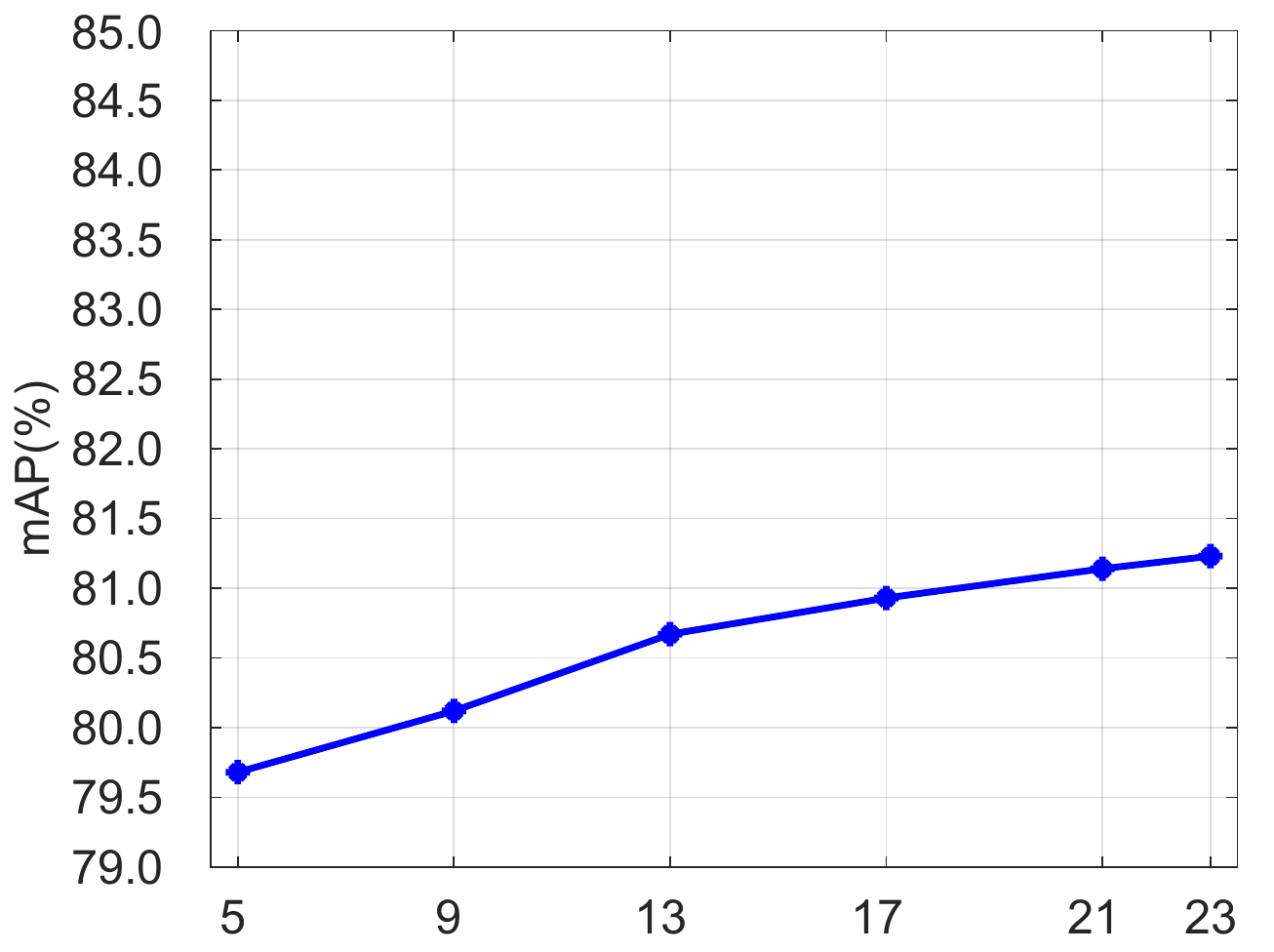}}
	\subfigure[\normalsize{(b) sampling strides} \label{fig:pic32}]{\includegraphics[height = 4.2cm, width = 0.315\linewidth]{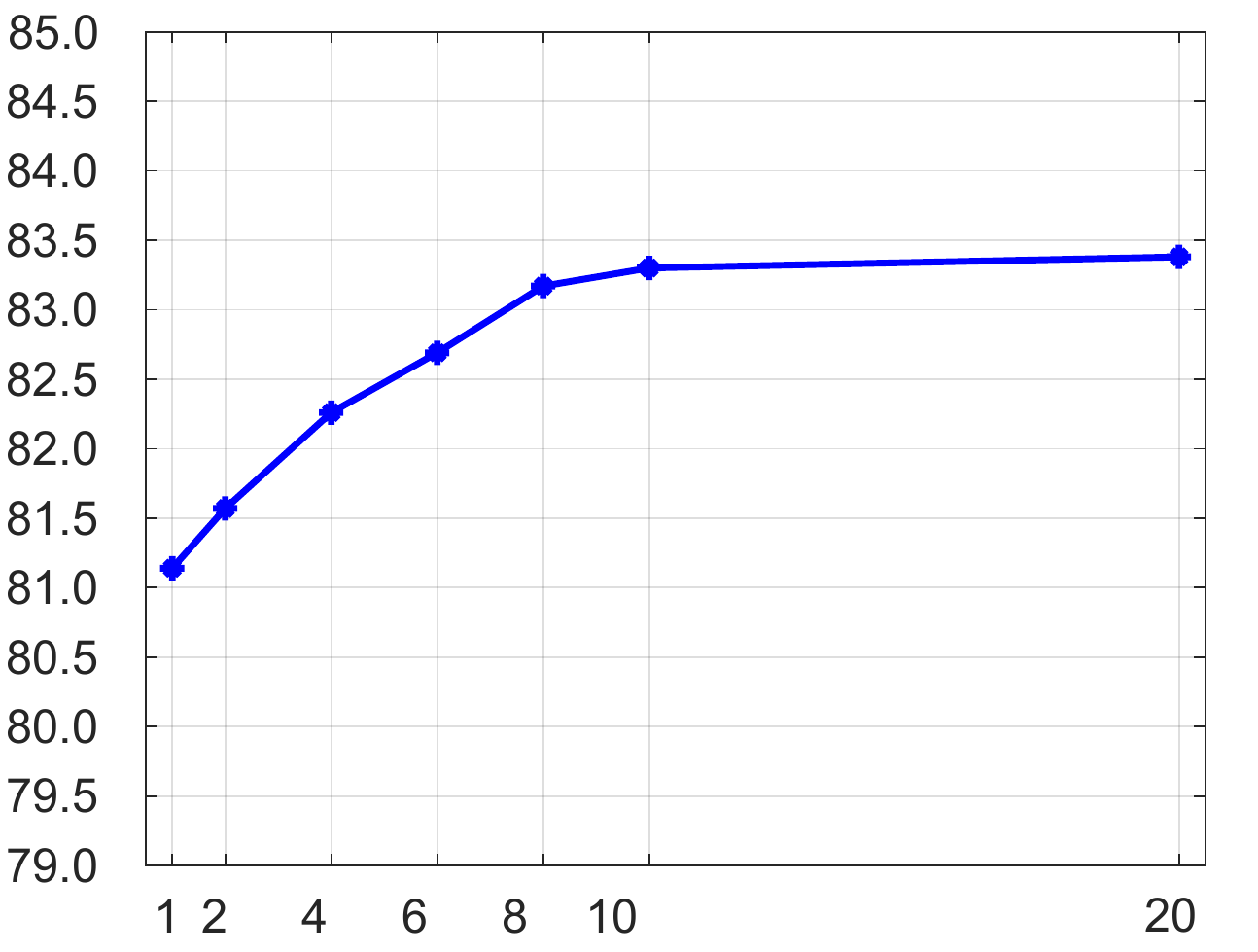}}
	\subfigure[\normalsize{(c) number of shuffled test frames} \label{fig:pic33}]{\includegraphics[height = 4.2cm, width = 0.315\linewidth]{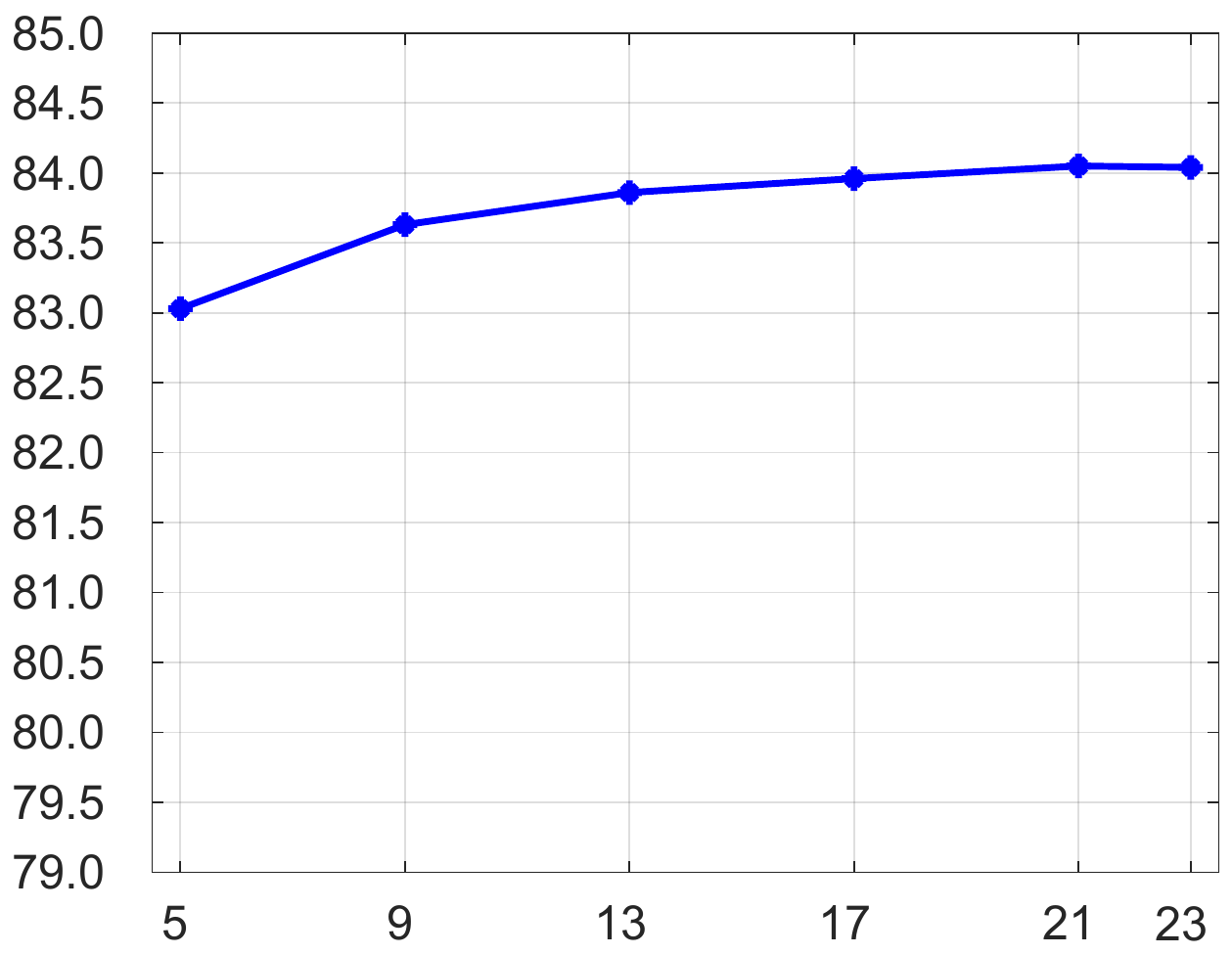}}
	\caption{The test performance on the ImageNet validation set obtained by the proposed DSFNet with (a) different numbers of test frames, (b) different sampling strides, and (c) different numbers of {\bfseries shuffled} test frames.}
	\label{fig:tb15}
\end{figure*}   

{\bfseries Dual Semantic Fusion.} We apply the frame-level fusion module to the output of $conv4$ to generate enhanced frame-level features. These enhanced features are then fed into RPN to get a series of object instances. For the instance-level fusion module, we insert it after the RoI pooling layer twice to generate enhanced instance-level features for final detection.

{\bfseries Training and Inference Details.} For both training and inference, the input images are resized to have a shorter side of 600 pixels. During training, the backbone feature extraction network (\emph{i.e.}, ResNet-101 or ResNeXt-101) is initialized with the weights that were pre-trained on the ImageNet classification dataset \cite{imagenet_cls}. The whole network is trained on 8 GPUs using SGD with cross-entropy loss. The total batch size is 8 with each GPU holding one sample. During training, a sample contains 3 frames: One is the current frame for training and the other two are the support frames that provide temporal information. For the ImageNet VID dataset, the two support frames are randomly sampled in the current video sequence. And for the ImageNet DET dataset, the three frames from the dataset (\emph{i.e.}, the still image dataset) are identical. We train the proposed network for a total of 247k iterations. The initial learning rate is set to $2.5 \times 10^{-4}$ and it is respectively dropped by a factor of 10 at the 109k and 219k iterations. In addition, we adopt the same data augmentation strategy as in \cite{SELSA}. During inference, we sample $n$ frames in a video for the proposed DSFNet. The influence of the parameter $n$ will be discussed in the next subsection.

\subsection{Frame Sampling Strategies}
Frame sampling is an essential part of feature enhancement based video object detection methods. This has been reported by the previous works (\emph{e.g.}, \cite{Fgfa,SELSA}). Therefore, it is worth investigating the effectiveness of DSFNet under different frame sampling strategies. 

Here, we evaluate the performance of the proposed DSFNet using a fixed interval sampling strategy with different numbers of test frames and various sampling strides. Moreover, we also use a stochastic sampling strategy to evaluate DSFNet. Let $n$ be the number of the test frames. The $n$ test frames consist of the evaluated frame and the $n-1$ sampled support frames.

Firstly, we evaluate the proposed DSFNet with different numbers of test frames using a fixed interval sampling strategy. During the evaluation, the $n-1$ adjacent frames are sampled as the support frames with a fixed sampling stride of 1. By increasing the number of the test frames from 5 to 23, 
the performance of DSFNet is improved from 79.7\% to 81.2\%  mAP (+1.5\%), as shown in Figure \ref{fig:pic31}. From the figure, it is clear that the performance of DSFNet can be improved with the increasing number of the test frames. However, more test frames require more computing resources. Therefore, more choices of the sampling stride should be considered before the value of $n$ (\emph{i.e.}, the number of test frames) is determined.

Then, we examine the influence of different sampling strides on the performance of DSFNet. Let $s$ be the sampling stride. The $n-1$ support frames are uniformly sampled at every $s$ frames. We fix the $n$ value to 21 but use various sampling strides on DSFNet. The experimental results are reported in Figure \ref{fig:pic32}. As we can see, the performance of DSFNet can be improved with the increasing of the sampling stride. In particular, DSFNet achieves the highest mAP of 83.4\% with the largest stride of 20, which is large enough to traverse most of the test video sequences in the ImageNet VID set. Actually, the current fixed interval sampling strategy can be considered as a special case of the stochastic sampling strategy, which randomly samples the support frames from the whole test video sequence.

Finally, we replace the fixed interval sampling strategy with the stochastic sampling strategy to further improve the performance of our DSFNet. The test frames are shuffled at the beginning. After that, we adjust the number of the shuffled test frames from 5 to 23 to evaluate the performance of DSFNet. The obtained results are given in Figure \ref{fig:pic33}, from which we can see that by leveraging the rich context information in the temporal domain, DSFNet achieves the highest mAP of 84.1\% when the number of the shuffled test frames is set to 21. Moreover, the performance of DSFNet is saturated when the number of the shuffled test frames is more than 21, as illustrated in Figure \ref{fig:pic33}. Consequently, we choose the stochastic sampling strategy to sample the test frames and fix the value of $n$ to 21 in the proposed DSFNet for all the following experiments, which is a good trade-off between effectiveness and efficiency.

\begin{figure*}[t]
	\centering
	\includegraphics[width = 0.90\linewidth]{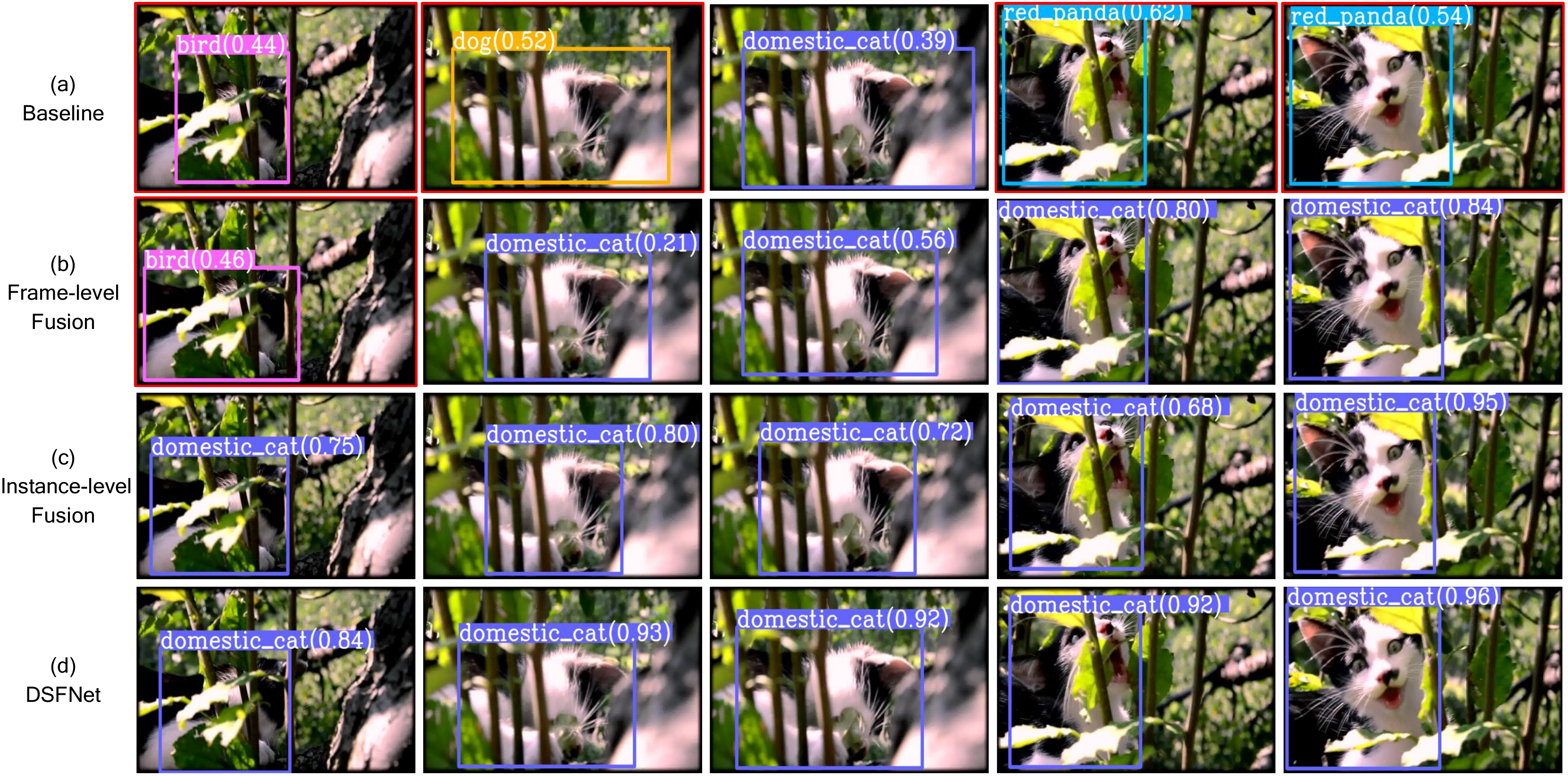}
	\caption{Qualitative results obtained by four variants of our DSFNet. The results include object labels and the corresponding confidence scores in brackets. The four variants of DSFNet are listed at the left of this figure. The detection results are marked in different colors according to their predicted labels. Those frames with false positive results are highlighted by red rectangles.}
	\label{fig:comparison}
\end{figure*}

\subsection{Ablation Study}
We perform several ablation studies on the ImageNet VID validation set to evaluate the effectiveness of the proposed DSFNet. Table \ref{tab:ablation} reports the quantitative results obtained by four variants of DSFNet, which are respectively: (a) the baseline, (b) the baseline with the proposed frame-level fusion module, (c) the baseline with the proposed instance-level fusion module, and (d) the proposed DSFNet. All the results in Table \ref{tab:ablation} are based on the ResNet-101 backbone. In particular, Table \ref{tab:ablation}(a) provides the results obtained by the baseline detector (\emph{i.e.}, Faster R-CNN).

\begin{table}[]
	\centering
	\caption{Ablation study on the ImageNet VID validation set. The results are obtained by four variants of DSFNet. The best results are highlighted by bold.}
	\newcommand{\tabincell}[2]{\begin{tabular}{@{}#1@{}}#2\end{tabular}}
	\begin{tabular}{c |c| c| c| c}
		\toprule
		&(a) &(b) &  (c)&  (d) \\
		\cline{2-5}
		Variants  & \tabincell{c}{ Baseline}& \tabincell{c}{ Frame-\\Level \\Fusion} &  \tabincell{c}{ Instance-\\Level\\Fusion} &\tabincell{c}{ DSFNet} \\
		\hline
		mAP (\%) & 74.7  & 77.0 & 83.3 & \multicolumn{1}{l}{$\bm{84.1_{\uparrow 9.4}}$} \\
		\hline
		mAP (\%) (slow) & 83.3 & 85.7 &89.6  &\multicolumn{1}{l}{$\bm{90.0_{\uparrow 6.7}}$} \\
		\hline
		mAP (\%) (medium) & 72.3 & 74.8  &81.4 & \multicolumn{1}{l}{$\bm{82.6_{\uparrow 10.3}}$} \\
		\hline
		mAP (\%) (fast) & 52.3 & 54.0  &66.2 & \multicolumn{1}{l}{$\bm{67.0_{\uparrow 14.7}}$} \\
		\bottomrule
	\end{tabular}
	\label{tab:ablation}
\end{table} 


{\bfseries Frame-level Fusion.} 
From the results in Table \ref{tab:ablation}(b), we can see that introducing the proposed frame-level fusion module into the baseline detector leads to $+2.3\%$ gain in terms of mAP. This is because that the proposed frame-level fusion module is capable of producing enhanced features by fusing the frame-level information. As a result, the frame-level fusion module can effectively propagate the beneficial semantic information across frames, by which it boosts the performance of the baseline detector.

{\bfseries Instance-level Fusion.}
Table \ref{tab:ablation}(c) shows the results obtained by applying the proposed instance-level feature fusion module to the baseline detector. Compared with the baseline detector, a significant $+8.6\%$ gain on mAP is achieved. This performance gain can be attributed to the improvement of fusing the rich semantic context information across instances and leveraging the proposed geometric similarity measure. The rich instance-level information makes the detector robust against object appearance variations (such as motion blur, occlusion, and deformation) in videos.

The overall performance of leveraging the above two fusion modules is presented in Table \ref{tab:ablation}(d). Jointly applying both of the proposed frame-level and instance-level feature fusion modules to the baseline detector leads to a considerable gain of $+9.4\%$ mAP on the test dataset. Moreover, as shown in Table \ref{tab:ablation}, DSFNet can significantly improve the detection performance of the baseline detector on all the three types of motion groups in \cite{Fgfa}. Specifically, DSFNet achieves $+6.7\%$, $+10.3\%$, and $+14.7\%$ mAP gains for the object detection on the slow, medium, and fast motion groups, respectively. The most significant improvement of $+14.7\%$ is achieved by DSFNet on the fast motion group. This is because that DSFNet can effectively enhance the deteriorated features of fast moving objects by fusing the features among frames and instances. Thus, the enhanced features contain beneficial semantic information from other high-quality frames and instances, which makes DSFNet more robust in dealing with the fast moving objects. Overall, the results in Table \ref{tab:ablation} show the effectiveness of combining both the frame-level and instance-level fusions in the proposed DSFNet within a unified framework for detecting objects in videos.

{\bfseries Qualitative Detection Results.}
Besides the quantitative results, we also provide some qualitative detection results obtained by these variants in Figure \ref{fig:comparison}. The video sequence in Figure \ref{fig:comparison} is very challenging due to the deteriorated appearance of the cat caused by serious occlusions and significant pose variations. As shown in Figure \ref{fig:comparison}(a), the baseline detector tends to classify the detected objects into incorrect categories. The frame-level fusion module utilizes the features from more than one frame and achieves much better results than the baseline detector using the features from a single frame. However, it fails in some hard cases (see the most left frame in Figure \ref{fig:comparison}(b)). Meanwhile, the detector with only the proposed instance-level fusion module can correctly detect those objects affected by occlusions and pose variations with relatively low confidence scores and less accurate bounding boxes, as shown in Figure \ref{fig:comparison}(c). Finally, by leveraging both the frame-level and instance-level fusion modules, the proposed DSFNet yields the best performance among those variants, which shows the effectiveness of DSFNet.

\begin{table}[t]
	\centering
	\caption{Comparison with state-of-the-art competitors on the ImageNet VID validation set. * indicates the methods with post-processing steps. The best results are highlighted by bold.}
	\begin{tabular}{c c c}
		\toprule
		Methods  & Backbone & mAP (\%) \\
		\hline
		FGFA \cite{Fgfa} & ResNet-101 & 76.3 \\
		MANet \cite{MANet} & ResNet-101 & 78.1 \\
		THP \cite{THP} & ResNet-101 & 78.6 \\
		STSN \cite{STSN} & ResNet-101 & 78.9 \\
		LRTR \cite{Leverage_long_range} & ResNet-101 & 81.0 \\
		RDN \cite{RDN} & ResNet-101 & 81.8 \\
		SELSA \cite{SELSA} & ResNet-101 & 82.7 \\
		\hline
		FGFA* \cite{Fgfa} & ResNet-101 & 78.4 \\
		ST-Lattice* \cite{ST-lattice} & ResNet-101 & 79.6 \\
		D\&T* \cite{DandT}  & ResNet-101 & 79.8 \\
		MANet* \cite{MANet} & ResNet-101 & 80.3 \\
		STSN* \cite{STSN}  & ResNet-101 & 80.4 \\
		STMN* \cite{STMN}  & ResNet-101 & 80.5 \\
		RDN* \cite{RDN} & ResNet-101 & 83.8\\
		\midrule
		DSFNet (ours)  & ResNet-101 & \bfseries{84.1} \\
		\midrule
        RDN \cite{RDN} & ResNeXt-101 & 83.2 \\
        LRTR \cite{Leverage_long_range} & ResNeXt-101 & 84.1 \\
		SELSA \cite{SELSA} & ResNext-101   & 84.3 \\
        \midrule
        FGFA* \cite{Fgfa} & Inception-ResNet & 80.1 \\
		D\&T* \cite{DandT}  & Inception-v4  & 82.1 \\
		RDN* \cite{RDN} & ResNeXt-101 & 84.7 \\
		\midrule
		DSFNet (ours)  & ResNeXt-101  &  $\bm{85.4}$ \\
		\bottomrule
		
	\end{tabular}
	\label{tab:comparison}
\end{table}

\subsection{Comparison with State-of-the-art Methods}
We compare the proposed DSFNet with several state-of-the-art video object detection methods, including MANet \cite{MANet}, FGFA \cite{Fgfa}, THP \cite{THP}, ST-Lattice \cite{ST-lattice}, D\&T \cite{DandT}, STSN \cite{STSN}, STMN \cite{STMN}, RDN \cite{RDN}, SELSA \cite{SELSA}, and LRTR \cite{Leverage_long_range}. Table \ref{tab:comparison} summarizes the results obtained by the proposed DSFNet and the other state-of-the-art methods on the ImageNet VID validation set. As shown in Table \ref{tab:comparison}, the proposed DSFNet with ResNet-101 obtains $84.1\%$ mAP, outperforming all the other competing video object detectors. 

Among these detectors, FGFA and THP propose to improve per-frame features by fusing the features across frames with external guidance using optical flow information estimated by \cite{flownet}. Thus, these two detectors may suffer from the instability of their guidance. In contrast, the proposed DSFNet aims to enhance the features for video object detection in a unified framework without using any external guidance, which yields much better performance than FGFA ($+7.8\%$ mAP) and THP ($+5.5\%$ mAP). In addition, STSN and STMN only use the aggregated frame-level features to perform robust video object detection. Compared with them, DSFNet achieves better results by fusing the frame-level and instance-level features, outperforming these two methods by $+5.2\%$ and $+3.6\%$ mAP, respectively. Meanwhile, SELSA is a newly proposed video object detection method that utilizes the appearance similarities among instances to perform instance-level semantic fusion. Compared with SELSA, DSFNet adopts both appearance similarity and geometric similarity in the instance-level semantic fusion module to mitigate the information distortion problem. As a result, DSFNet achieves the highest mAP of 84.1\%, which outperforms SELSA by $+1.4\%$ mAP. RDN also aggregates the instance-level features across frames to generate the enhanced instance-level features for robust detection, and it achieves a satisfying performance of $81.8\%$ mAP. Moreover, RDN employs additional post-processing techniques to boost its performance from $81.8\%$ to $83.8\%$ mAP. Nevertheless, the performance of RDN is still inferior to that of the proposed DSFNet, which does not use any post-processing techniques.

Moreover, by changing the backbone feature extractor from ResNet-101 to a stronger backbone feature extractor ResNeXt-101, our DSFNet achieves a better performance of $85.4\%$ mAP without using any post-processing steps. This result still outperforms the reported results from the current state-of-the-art video object detection methods that use stronger backbone networks, as shown in Table \ref{tab:comparison}. The $+1.3\%$ performance gain on mAP achieved by the ResNeXt version of DSFNet can be ascribed to the more powerful features extracted by the stronger backbone network. As a result, the fused features in the ResNeXt version of DSFNet contain more beneficial semantic information, which makes it more robust in handling the aforementioned challenges 
in video object detection.

\section{Conclusion}
\label{sec:Conclusion}
In this paper, we present a novel dual semantic fusion network (named DSFNet) for video object detection. In DSFNet, both frame-level and instance-level semantics contained in input videos are distilled and fused to generate enhanced features for robust video object detection. Different from the existing one-stage feature enhancement methods that perform the feature fusion at either frame level or instance level with external guidance, DSFNet combines both frame-level and instance-level feature fusions, which can be learned in a unified fusion framework without any external guidance. In addition, we also introduce a new geometric similarity measure to mitigate the information distortion caused by noise during the fusion process. Extensive experiments on the large scale ImageNet VID dataset demonstrate the effectiveness and superiority of the proposed DSFNet. In particular, compared with several other cutting-edge methods, DSFNet has achieved the best performance of 84.1\% mAP with ResNet-101 and 85.4\% mAP with ResNeXt-101 without using any post-processing steps. Moreover, the proposed two-stage semantic fusion scheme in DSFNet is generic for video object detection, which can inspire more future works.

\begin{acks}
This work is supported by the National Science Foundation of China under Grant U1605252 and Grant 61872307.
\end{acks}


\bibliographystyle{ACM-Reference-Format}
\bibliography{egbib}

\end{document}